\documentclass[11pt]{article}
\usepackage{coling2018}
\usepackage{times}
\usepackage{url}
\usepackage{latexsym}

\usepackage{graphicx}
\usepackage{xspace}
\usepackage{amsmath}
\usepackage{amsfonts}
\usepackage{mathrsfs}
\usepackage{array,multirow}
\usepackage{pgfplots}
\usepackage{tikz}
\usepackage{subfigure}
\usepackage{array} 
\usepackage{underscore}
\usepackage{verbatim}
\usepackage{ulem}
\usepackage{arydshln}
\usepackage{color}

\usetikzlibrary{backgrounds,fit}
\usetikzlibrary{shapes,arrows,shadows}
\usetikzlibrary{patterns}
\usetikzlibrary{shapes.geometric}
\usetikzlibrary{decorations.pathreplacing}
\usetikzlibrary{calc}
\usepackage{ulem}
\usepackage{arydshln}
\usepackage{color}
\usepackage{caption}

\newcommand{\PreserveBackslash}[1]{\let\temp=\\#1\let\\=\temp}
\newcolumntype{C}[1]{>{\PreserveBackslash\centering}p{#1}}
\newcolumntype{R}[1]{>{\PreserveBackslash\raggedleft}p{#1}}
\newcolumntype{L}[1]{>{\PreserveBackslash\raggedright}p{#1}}

\newlength{\vseg}
\newlength{\hseg}
\newlength{\wnode}
\newlength{\hnode}
\definecolor{ugreen}{rgb}{0,0.5,0}
\definecolor{lgreen}{rgb}{0.9,1,0.8}
\definecolor{lightgray}{gray}{0.85}
\usepackage{xcolor}
\definecolor{myblack}{rgb}{0.15,0.15,0.15}

\usepackage{collcell}
\newcommand*{\MinNumber}{0}%
\newcommand*{\MaxNumber}{1}%
\newcommand{\ApplyGradient}[1]{%
	\pgfmathsetmacro{\PercentColor}{100.0*(#1-\MinNumber)/(\MaxNumber-\MinNumber)}
	\hspace{-0.33em}\colorbox{white!\PercentColor!myblack}{}
}
\newcolumntype{Q}{>{\collectcell\ApplyGradient}c<{\endcollectcell}}

\makeatletter
\def\py@yunpriv#1{%
	\if a#1 10\else
	\if o#1 9\else
	\if e#1 8\else
	\if i#1 7\else
	\if u#1 6\else
	\if v#1 5\else
	\if A#1 4\else
	\if O#1 3\else
	\if E#1 2\fi\fi\fi\fi\fi\fi\fi\fi\fi0
}

\def\py@init{%
	\edef\py@befirst{}%
	\edef\py@char{}\edef\py@tuneletter{}%
	\def\py@last{}%
	\def\py@tune{5}%
}

\def\pinyin#1{%
	\edef\py@postscan{#1}%
	\py@init
	\loop
	\edef\py@char{\expandafter\@car\py@postscan\@nil}%
	\edef\py@postscan{\expandafter\@cdr\py@postscan\@nil}%
	\ifnum 0 < 0\py@char
	\edef\py@tune{\py@char}%
	\py@first \py@tuneat\py@tuneletter\py@tune \py@last\kern -4sp\kern 4sp{}\py@init
	\else
	\ifnum\py@yunpriv\py@char > \py@yunpriv\py@tuneletter
	\edef\py@tuneletter{\py@char}\edef\py@first{\py@befirst}\def\py@last{}%
	\else
	\edef\py@last{\py@last\if v\py@char\"u\else\py@char\fi}%
	\fi
	\edef\py@befirst{\py@befirst\if v\py@char\"u\else\py@char\fi}%
	\fi
	\ifx\py@postscan\@empty\else
	\repeat
}

\let\py@macron \=
\let\py@acute \'
\let\py@hacek \v
\let\py@grave \`

\def\py@tuneat#1#2{%
	\if v#1%
	\py@tune@v #2%
	\else
	\if i#1%
	\py@tune@i #2%
	\else
	\ifcase#2%
	\or\py@macron #1\or\py@acute #1\or\py@hacek #1\or\py@grave #1\else #1%
	\fi
	\fi\fi
}

\def\py@tune@v#1{{%
		\dimen@ii 1ex%
		\fontdimen5\font 1.1ex%
		\rlap{\"u}%
		\fontdimen5\font .6ex%
		\ifcase#1%
		\or\py@macron u\or\py@acute u\or\py@hacek u\or\py@grave u\else u%
		\fi
		\fontdimen5\font\dimen@ii
}}

\def\py@tune@i#1{%
	\ifcase#1
	\or\py@macron \i\or\py@acute \i\or\py@hacek \i\or\py@grave \i\else i%
	\fi
}
\makeatletter

\title{Multi-layer Representation Fusion for Neural Machine Translation}

\author{
	Qiang Wang$^{\dagger}$,
	Fuxue Li$^{\dagger,\ddagger}$,
	Tong Xiao$^{\dagger*}$,
	Yanyang Li$^{\dagger}$,
	Yinqiao Li$^{\dagger}$,
	Jingbo Zhu$^{\dagger}$ \\
	$^{\dagger}$Natural Language Processing Lab., Northeastern University \\
	$^{\ddagger}$YingKou Institute of Technology \\
	{\tt
		wangqiangneu@gmail.com,
		lifuxue119@163.com}\\
	{\tt
		xiaotong@mail.neu.edu.cn,
		blamedrlee@outlook.com} \\
	{\tt 
		li.yin.qiao.2012@hotmail.com,
		zhujingbo@mail.neu.edu.cn} \\
	}

\date{}

\begin{document}
	\maketitle
	\begin{abstract}
		Neural machine translation systems require a number of stacked layers for deep models. But the prediction depends on the sentence representation of the top-most layer with no access to low-level representations. This makes it more difficult to train the model and poses a risk of information loss to prediction. In this paper, we propose a multi-layer representation fusion (MLRF) approach to fusing stacked layers. In particular, we design three fusion functions to learn a better representation from the stack. Experimental results show that our approach yields improvements of 0.92 and 0.56 BLEU points over the strong Transformer baseline on IWSLT German-English and NIST Chinese-English MT tasks respectively. The result is new state-of-the-art in German-English translation.
		
	\end{abstract}
	
	\section{Introduction}
	\label{sec:introduction}
	
	\blfootnote{* Corresponding author} 
	
	%
	%
	\blfootnote{
	%
	%
	%
	%
	%
	%
	 \hspace{-0.65cm}  
	 This work is licensed under a Creative Commons 
	 Attribution 4.0 International License.
	 License details:
	 \url{http://creativecommons.org/licenses/by/4.0/}
	}
	
	\noindent Neural models that use the encoder-decoder architecture to capture the translation equivalence relation between languages have been widely adopted over the last few years. The simplest of these relies on one recurrent neural network layer on both the encoder and decoder sides \cite{bahdanau2014neural}, whereas others have successfully explored the high-level representation of language via deeper models \cite{wu2016google,gehring2017convs2s,vaswani2017attention}. It has been noted that increasing the network depth is one of the factors contributing to the success of neural machine translation (NMT). To this end, one can stack a number of layers for an enriched sentence representation. E.g., popular NMT systems require 4 stacked layers or more for state-of-the-art results on large-scale translation tasks \cite{wu2016google}. Unfortunately, a straightforward implementation of deep neural networks has been found to underperform shallow models due to the poor convergence. One solution to this problem is to add identity mapping (or shortcut connection) between layers. A popular method is the residual network \cite{he2016deep}, which has been used in several successful systems, e.g., GNMT and Transformer.
	
	This model offers a good way to ease the information flow in the stack. But they do not make full use of the multi-level representations of the deep networks. E.g., word predictions depend on the sentence representation of the top-most layer with no access to low-level representations. Recent evidence supports that it helps when features from lower-level layers are involved in prediction \cite{xiong2017multi,gehring2017convs2s}. A good instance is DenseNet that applies a fully-connected convolutional network to computer vision \cite{huang2017densely}. But it is still rare to see studies on this issue in machine translation.
	
	In this paper, we take a further step along the line of this research. Drawing on previous successful attempts to create shortcut connections between adjacent layers, we propose a multi-layer representation fusion (MLRF) approach that connects one layer to all its predecessors. It learns a fused representation by accessing all lower-level representations of sentence, rather than learning from the limited information flow in one residual connection. The contributions of this work are three-fold.
	
	\begin{itemize}
		\item We propose an approach to learning better sentence representations by accessing lower-level layers. To our knowledge, it is the first time to use densely connected networks in NMT.
		\item We investigate methods to learn functions that fuse different levels of sentence representations. In particular, we propose to learn a 2D representation by the self-attention mechanism.
		\item We demonstrate the effectiveness of the approach on top of a strong system (Transformer) which has already used residual networks. It achieves a new state-of-the-art result on the IWSLT German-English MT task.
	\end{itemize}
	
	Our approach is applicable to arbitrary neural architectures and is easy to be implemented. In experiments on IWSLT German-English and NIST Chinese-English translation, it yields improvements of 0.92 and 0.56 BLEU points over the baseline respectively. More interestingly, we find that our approach has a regularizing effect and reduces the risk of over-fitting.
	
	\section{The Transformer System}
	\label{sec:transformer}
	
	In this work, all discussions and experiments are based on the Transformer system. We choose Transformer because it is one of the most successful NMT systems in recent MT evaluations. Unlike usual NMT models, Transformer does not require any recurrent units for modeling word sequences of arbitrary length. Instead, it resorts to self-attention and standard feed-forward networks for both encoder and decoder.
	
	On the encoder side, there are $L$ identical stacked layers. Each of them is composed of a self-attention sub-layer and a feed-forward sub-layer. The attention model used in Transformer is scaled dot-product attention\footnote{Given a sequence of vectors and a position $i$, the self-attention model computes the dot-product of the input vectors for each pair of positions $(i,j)$, followed by a rescaling operation and Softmax. In this way, we have an attention score (or weight) for each $(i, j)$. It is then used to generate the output by a weighted sum over all input vectors.}. Its output is fed into a fully connected feed-forward network. To ease training, the output of each sub-layer is defined as $\textrm{LayerNorm}(x + \textrm{sublayer}(x))$, where $\textrm{LayerNorm}(\cdot)$ is layer normalization \cite{ba16layer} and $\textrm{sublayer}(x)$ is the output of the sub-layer. The identical mapping of input $x$ represents the residual connection. To facilitate description, we use $H=\{h^1, \ldots, h^L\}$ to denote the outputs of source-side layers in this paper \footnote{We regard the word embedding layer as a special layer at the bottom of the stack, and denote it as $h^0$.}.
	
	Likewise, the decoder has another stack of $L$ identical layers (denoted as the function $\textrm{Layer}(\cdot)$). It has an encoder-decoder attention sub-layer in addition to the two sub-layers used in each encoder layer. Moreover, because the model is auto-regressive, the decoder attends a target position to all positions up to it. Let $z_j^l$ be the output vector of target position $j$ in the $l$-th layer, and $z_{\le j}^{l-1}$ be the vector sequence $\{z_{1}^{l-1}, \ldots, z_{j}^{l-1}\}$. The output of the layer can be described as $z_j^l=\textrm{Layer}(z_{\le j}^{l-1}, H)$.
	
	The auto-regressive property also persists in the output layer. Given a source sentence $\textbf{x}=(x_1,\ldots,x_{M})$, and a target sentence $\textbf{y}=(y_1,\ldots,y_{N})$, the translation model is defined to be:
	
	\begin{equation}
	P(\textbf{y}|\textbf{x})=\prod_{j=1}^{N}{\Pr(y_j|y_{<j},\textbf{x})}
	\end{equation}
	
	\noindent where $\Pr(y_j|y_{<j},\textbf{x})$ is the probability of generating a target word $y_j$ given the previously generated words $y_{<j}$ and the source sentence $\textbf{x}$. To model $\Pr(y_j|y_{<j},\textbf{x})$, we have
	
	\begin{equation}
	\label{eq:core}
	\Pr(y_j|y_{<j},\textbf{x})=\textrm{Softmax}(W_o \cdot z_j^L + b_o)
	\end{equation}
	
	\noindent where $W_o$ and $b_o$ are the model parameters of the output layer. Obviously, the word prediction model here depends only on the output of the top-most layer in the stack (i.e., $z_j^L$).


	
	
	
	\section{The Approach}
	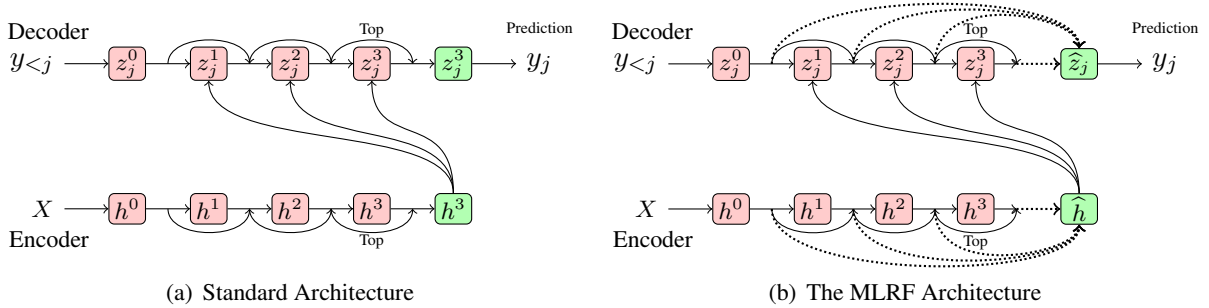
\begin{figure*}[t]
		\begin{center}
			\setlength{\tabcolsep}{1pt}
			\begin{tabular}{cc}
				\subfigure [Standard Architecture]
				{
					\begin{tikzpicture}
					\begin{scope}
					
					\setlength{\vseg}{1.5em}
					\setlength{\hseg}{2.5em}
					\setlength{\wnode}{2.5em}
					\setlength{\hnode}{1.5em}
					\tikzstyle{wordnode} = [thick,minimum height=0.7\hnode,minimum width=0.5\wnode]
					\tikzstyle{embnode} = [draw, thin, rounded corners=2pt, inner sep=1pt, fill=yellow!30, minimum height=0.7\hnode, minimum width=0.5\wnode]
					\tikzstyle{hiddennode} = [draw, thin, rounded corners=2pt, inner sep=1pt, fill=red!20, minimum height=0.7\hnode, minimum width=0.5\wnode]
					\tikzstyle{probnode} = [fill=blue!30,minimum width=0.2\wnode]
					\tikzstyle{outnode} = [draw, thin, rounded corners=2pt, inner sep=1pt, fill=green!30, minimum height=0.7\hnode, minimum width=0.5\wnode]
					
					\node [] (start) at (0,0) {};
					\node [wordnode] (xseq) at ([yshift=0.9\hseg]start) {\footnotesize{$X$}};
					\node [hiddennode, anchor=west] (h0) at ([xshift=\vseg] xseq.east) {\footnotesize{$h^0$}};
					
					\node [wordnode, anchor=east] (yseq) at ([yshift=2\hseg] xseq.east) {$y_{<j}$};
					\node [hiddennode, anchor=west] (z0) at ([xshift=\vseg] yseq.east) {\footnotesize{$z_j^0$}};
					
					\foreach \x/\y/\b in {1/1/0,2/2/1,3/3/2}
					{
						\node [hiddennode, anchor=west] (h\x) at ([xshift=\vseg] h\b.east) {\footnotesize{$h^\y$} };					
						\node [hiddennode, anchor=west] (z\x) at ([xshift=\vseg] z\b.east) {\footnotesize{$z_j^\y$} };
					}
					
					\foreach \x in {0,1,2,3}
					{
						\coordinate (lh\x) at ([xshift=-0.5\vseg] h\x.west);
						\coordinate (rh\x) at ([xshift=0.5\vseg] h\x.east);
						\coordinate (lz\x) at ([xshift=-0.5\vseg] z\x.west);
						\coordinate (rz\x) at ([xshift=0.5\vseg] z\x.east);
					}
					
					\draw [->,thin] (xseq.east) -- ([yshift=0pt] h0.west);
					\draw [->,thin] (yseq.east) -- ([yshift=0pt] z0.west);
					
					\foreach \x/\y in {0/1,1/2,2/3}
					{
						\draw [->,thin] (h\x.east) -- ([yshift=0pt] h\y.west);
						\draw [->,thin] (z\x.east) -- ([yshift=0pt] z\y.west);		
						
						\draw [->, thin] (lh\y) to [out=-90,in=-90] (rh\y);
						\draw [->, thin] (lz\y) to [out=90,in=90] (rz\y);
					}
					
					\node [outnode, anchor=west] (encout) at ([xshift=\vseg] h3.east) {\footnotesize$h^3$};
					\draw [->,thin] (h3.east) -- (encout.west);
					
					\node [outnode, anchor=west] (decout) at ([xshift=\vseg] z3.east) {\footnotesize$z_j^3$};
					\draw [->,thin] (z3.east) -- (decout.west);
					\node [anchor=west] (decoutput) at ([xshift=\vseg] decout.east) {$y_j$};
					\draw [->,thin] (decout.east) -- (decoutput.west);
					
					\foreach \x/\y in {1/0,2/1,3/2}
					{
						\draw [->,thin] (encout.north).. controls +(north:0.1*\y*\hseg+\hseg) and +(south:\hseg) ..(z\x.south);
					}
					
					\node [anchor=north east] (enclabel) at ([xshift=0.8\vseg,yshift=0.1\hseg]xseq.south east) {\small{Encoder}};
					\node [anchor=south east] (declabel) at ([xshift=0.8\vseg,yshift=-0.1\hseg]yseq.north east) {\small{Decoder}};
					\node [anchor=north] (enctop) at ([yshift=0\hseg]h3.south) {\tiny{Top}};
					\node [anchor=south] (dectop) at ([yshift=-0.015\hseg]z3.north) {\tiny{Top}};
					\node [anchor=south] (pred) at ([yshift=0\hseg]decoutput.north) {\tiny{Prediction}};

					\end{scope}
					\label{fig:standar-nmt}
					\end{tikzpicture}
				}
				&
				\subfigure [The MLRF Architecture]
				{
					\begin{tikzpicture}[
					triangle/.style = {fill=blue!20, regular polygon, regular polygon sides=3, inner sep=0.5pt},
					]
					\begin{scope}
					
					\setlength{\vseg}{1.5em}
					\setlength{\hseg}{2.5em}
					\setlength{\wnode}{2.5em}
					\setlength{\hnode}{1.5em}
					\tikzstyle{wordnode} = [thick,minimum height=0.7\hnode,minimum width=0.5\wnode]
					\tikzstyle{embnode} = [draw, thin, rounded corners=2pt, inner sep=1pt, fill=yellow!30, minimum height=0.7\hnode, minimum width=0.5\wnode]
					\tikzstyle{hiddennode} = [draw, thin, rounded corners=2pt, inner sep=1pt, fill=red!20, minimum height=0.7\hnode, minimum width=0.5\wnode]
					\tikzstyle{outnode} = [draw, thin, rounded corners=2pt, inner sep=1pt, fill=green!30, minimum height=0.7\hnode, minimum width=0.5\wnode]
					\tikzstyle{probnode} = [fill=blue!30,minimum width=0.2\wnode]
					
					\node [wordnode] (xseq) at (0,0) {\footnotesize{$X$}};
					\node [hiddennode, anchor=west] (h0) at ([xshift=\vseg] xseq.east) {\footnotesize{$h^0$}};
					
					\node [wordnode, anchor=east] (yseq) at ([yshift=2\hseg] xseq.east) {$y_{<j}$};
					\node [hiddennode, anchor=west] (z0) at ([xshift=\vseg] yseq.east) {\footnotesize{$z_j^0$}};
					
					\foreach \x/\y/\b in {1/1/0,2/2/1,3/3/2}
					{
						\node [hiddennode, anchor=west] (h\x) at ([xshift=\vseg] h\b.east) {\footnotesize{$h^\y$} };					
						\node [hiddennode, anchor=west] (z\x) at ([xshift=\vseg] z\b.east) {\footnotesize{$z_j^\y$} };
					}
					
					\foreach \x in {0,1,2,3}
					{
						\coordinate (lh\x) at ([xshift=-0.5\vseg] h\x.west);
						\coordinate (rh\x) at ([xshift=0.5\vseg] h\x.east);
						\coordinate (lz\x) at ([xshift=-0.5\vseg] z\x.west);
						\coordinate (rz\x) at ([xshift=0.5\vseg] z\x.east);
					}
					
					\draw [->,thin] (xseq.east) -- ([yshift=0pt] h0.west);
					\draw [->,thin] (yseq.east) -- ([yshift=0pt] z0.west);
					
					\foreach \x/\y in {0/1,1/2,2/3}
					{
						\draw [->,thin] (h\x.east) -- ([yshift=0pt] h\y.west);
						\draw [->,thin] (z\x.east) -- ([yshift=0pt] z\y.west);		
						
						\draw [->, thin] (lh\y) to [out=-90,in=-90] (rh\y);
						\draw [->, thin] (lz\y) to [out=90,in=90] (rz\y);
					}
					
					
					\node [outnode, anchor=west] (encfusion) at ([xshift=1.5\vseg] h3.east) {\footnotesize$\widehat{h}$};
					\draw [->,thin] (h3.east) -- (rh3);
					\draw [->,thick,densely dotted] (rh3) -- (encfusion.west);
					
					\foreach \x/\y in {0/3,1/2,2/1}
					{
						\draw [->,thick,densely dotted] (rh\x).. controls +(south:\hseg) and +(south:0.2*\y*\hseg+0.2\hseg) ..(encfusion.south);
					}
					
					\node [outnode,anchor=west] (decfusion) at ([xshift=1.5\vseg] z3.east) {\footnotesize$\widehat{z}_j$};
					\draw [->,thin] (z3.east) -- (rz3);
					\draw [->,thick,densely dotted] (rz3) -- (decfusion.west);
					
					\node [anchor=west] (decoutput) at ([xshift=\vseg] decfusion.east) {$y_j$};
					\draw [->,thin] (decfusion.east) -- (decoutput.west);
					
					\foreach \x/\y in {0/3,1/2,2/1}
					{
						\draw [->,thick,densely dotted] (rz\x).. controls +(north:1.05\hseg) and +(north:0.2*\y*\hseg+0.25\hseg) ..(decfusion.north);
					}
					
					\foreach \x/\y in {1/0,2/1,3/2}
					{
						\draw [->,thin] (encfusion.north).. controls +(north:0.1*\y*\hseg+\hseg) and +(south:\hseg) ..(z\x.south);
					}
					
					\node [anchor=north east] (enclabel) at ([xshift=0.8\vseg,yshift=0.1\hseg]xseq.south east) {\small{Encoder}};
					\node [anchor=south east] (declabel) at ([xshift=0.8\vseg,yshift=-0.1\hseg]yseq.north east) {\small{Decoder}};
					\node [anchor=north] (enctop) at ([yshift=0\hseg]h3.south) {\tiny{Top}};
					\node [anchor=south] (dectop) at ([yshift=-0.015\hseg]z3.north) {\tiny{Top}};
					\node [anchor=south] (pred) at ([yshift=0\hseg]decoutput.north) {\tiny{Prediction}};
					
					\end{scope}
					\label{fig:mlrf-nmt}
					\end{tikzpicture}
				}
				
			\end{tabular}
		\end{center}
		
		\begin{center}
			\vspace{-0.5em}
			\caption{ Network Architectures. Dotted lines denote multi-layer fusion connections. Green rectangles denote the representations used in the downstream components.}
			\label{fig:methods}
			\vspace{-1.0em}
		\end{center}
	\end{figure*}
		
	The model presented in Section \ref{sec:transformer} shows a standard way of word prediction where the output layer is connected to the top-most hidden layer only. This is fine if the system is composed of one hidden layer or two. But it may be problematic when the network goes deeper. First, this method has difficulties in training deep models. To learn model parameters, gradients have to be propagated through a single channel of several stacked layers. Previous work has pointed out that it is not easy for lower-level layers to find good parameters due to the vanished or exploded gradients \cite{srivastava2015highway,he2016deep}. The situation is even worse when we use many non-linear transformations that are hard to learn. The residual connection between adjacent layers can alleviate the problem, but it is still a long distance for bottom layers to have direct feedbacks from the prediction. Second, there is potential information loss when we feed a single layer to the output layer. Top-level layers may forget previous features, especially when the stacked layers have the same capacity (e.g., the same size of layer output). For a simple solution, one can enlarge the layer size on the top. But it in turn drastically increases the number of parameters and is obviously not efficient for practical systems.
	
	In this work, we propose a multi-layer representation fusion method to address these problems. It learns a fused representation of the sentence by direct access to all the layers in the stack. In this way, the prediction can benefit from the fused representation which has less information loss. As another bonus, this method makes it easier for the deep network to learn parameters because of the efficient information flow in connections from bottom to top.
	
	\subsection{Multi-layer Representation Fusion}
	
	The idea of representation fusion is pretty simple. We introduce a fusion layer into the network, and connect it to all the stacked layers. This layer can learn a refined representation from different levels of the stack. Let $Z=\{z^1, \ldots, z^L\}$ be the output sequence of the stacked layers on the decoder side. We define $\phi(Z)$ to be the fusion function that fuses $\{z^1, \ldots, z^L\}$ into a single representation. More specifically, given a target word position $j$, $\widehat{z}_j=\phi(Z_j)$ is defined to be the fused representation of all layer representations $Z_j=\{z_j^1, \ldots, z_j^L\}$ in the stack for position $j$. Then Eq. (\ref{eq:core}) can be rewritten as:
	
	\begin{equation}
	\label{eq:fusion-prediction}
	\Pr(y_j|y_{<j},\textbf{x})=\textrm{Softmax}(W_o \cdot \phi(Z_j) + b_o)
	\end{equation}
	
	There are many choices to design the fusion function $\phi(Z_j)$. E.g., we can cast the usual method as a special case of our model and do exactly the same thing as in Section \ref{sec:transformer}, i.e., $\phi(Z_j)=z_j^L$. Alternatively, we can fuse more layers with another neural network. Moreover, we can do representation fusion on either the encoder side, the decoder side, or both. See Figure \ref{fig:methods} for an illustration of the method.
	
	Another note on representation fusion. The method is applicable to arbitrary neural network architectures with multiple levels of layers. Although we restrict ourselves to Transformer for our experiments, Eq. (\ref{eq:fusion-prediction}) actually shows a more general case. In a sense, our model is doing something similar to fully connected networks in computer vision \cite{huang2017densely,liu2018path}. Representation fusion is essentially a process that creates a densely connected network on top of the stack. In this work, we describe the problem in the more general framework, and will show that NMT systems can benefit from better design of fusion functions.

	\subsection{Fusion Functions}
	\label{sec:TWSA}
	
	\begin{figure*}[t]
		\begin{center}
			\begin{tikzpicture}[decoration=brace]
			\begin{scope}
			
			\setlength{\vseg}{1.5em}
			\setlength{\hseg}{2.5em}
			\setlength{\wnode}{3em}
			\setlength{\hnode}{2em}
			\tikzstyle{wordnode} = [thick,minimum height=0.7\hnode,minimum width=0.5\wnode]
			\tikzstyle{embnode} = [draw, thin, rounded corners=2pt, inner sep=1pt, fill=yellow!30, minimum height=0.7\hnode, minimum width=0.5\wnode]
			\tikzstyle{hiddennode} = [draw, thin, rounded corners=2pt, inner sep=1pt, fill=red!20, minimum height=0.7\hnode, minimum width=0.5\wnode]
			\tikzstyle{outnode} = [draw, thin, rounded corners=2pt, inner sep=1pt, fill=green!30, minimum height=0.7\hnode, minimum width=0.5\wnode]
			\tikzstyle{probnode} = [fill=blue!30,minimum width=0.2\wnode]
			
			\tikzstyle{weightnode} = [draw, thin, inner sep=1pt, minimum height=0.5\hnode, minimum width=\wnode]
			\tikzstyle{representnode} = [draw, thin, rounded corners=1pt, inner sep=1pt, minimum height=0.5\hnode, minimum width=\wnode]
			\tikzstyle{sumnode} = [draw, thin, rounded corners=1pt, inner sep=1pt, minimum width=2\wnode, minimum height=0.4\hnode]
			\tikzstyle{layernode} = [draw, thin, rounded corners=1pt, inner sep=1pt]
			
			\coordinate (h0) at (0,0);
			\foreach \x/\y in {1/0,2/1,3/2}
			{
				\node[representnode, anchor=north] (h\x) at ([yshift=\hseg]h\y.south) {\tiny$z_j^\x$};
			}
			\draw[decorate] (h1.south east) to node [auto] {\tiny$d$} (h1.south west);
			\foreach \x in {1,2,3}
			{
				\node[representnode, anchor=west] (i\x) at ([xshift=0.8\vseg]h\x.east) {};
			}
			\draw[decorate] (i1.south east) to node [auto] {\tiny$d_a$} (i1.south west);
			\foreach \x in {1,2,3}
			{
				\coordinate (p\x0) at ([xshift=0.8\vseg]i\x.south east);
				\foreach \y/\z/\c in {1/0/purple,2/1/yellow,3/2/red,4/3/blue,5/4/green}
				{
					\node[representnode, anchor=south west, minimum width=0.25\wnode, fill=\c!30] (p\x\y) at (p\x\z.south east) {};
				}
			}
			
			\node [inner sep=1pt] at (p15) {\tiny{$e_5^1$}};
			\node [inner sep=1pt] at (p25) {\tiny{$e_5^2$}};
			\node [inner sep=1pt] at (p35) {\tiny{$e_5^3$}};
			\draw[decorate, thin](p15.south east) to node [auto] {\tiny$n_{hop}$} (p11.south west);
			\foreach \x in {1,2,3}
			{
				\coordinate (a\x0) at ([xshift=1.5\vseg]p\x5.south east);
				\foreach \y/\z/\c in {1/0/purple,2/1/yellow,3/2/red,4/3/blue,5/4/green}
				{
					\node[representnode, anchor=south west, minimum width=0.25\wnode, fill=\c!30] (a\x\y) at (a\x\z.south east) {};
				}
			}
			\node[anchor=center] (num1) at (a15.center) {\tiny$.5$};
			\node[anchor=center] (num2) at (a25.center) {\tiny$.2$};
			\node[anchor=center] (num3) at (a35.center) {\tiny$.3$};
			\node[weightnode, anchor=south] (w1) at ([yshift=0.4\hseg]i3.north) {\small$W_1$};
			\draw[decorate] (w1.north west) to node [auto] {\tiny$d_a$} (w1.north east);
			\draw[decorate] (w1.south west) to node [auto] {\tiny$d$} (w1.north west);

			\node[weightnode, anchor=south] (w2) at ([yshift=0.4\hseg]p33.north) {\small$W_2$};
			\draw[decorate] (w2.north west) to node [auto] {\tiny$n_{hop}$} (w2.north east);
			\draw[decorate] (w2.south west) to node [auto] {\tiny$d_a$} (w2.north west);
			
			\coordinate (s0) at ([xshift=3.5\vseg, yshift=0.7\hseg]a24.east);
			\foreach \x/\y/\c in {1/0/purple,2/1/yellow,3/2/red,4/3/blue,5/4/green}
			{
				\node [sumnode, anchor=north west, fill=\c!30] (s\x) at ([xshift=-0.3\vseg, yshift=-0.15\hseg]s\y.north west) {};
			}
			\node[anchor=center] () at (s5.center) {\tiny$\tilde{s}_5$};
			\draw[decorate] (s5.south east) to node [auto] {\tiny$d$} (s5.south west);
			\draw[decorate] (s1.south east) to node [auto] {\tiny$n_{hop}$} (s5.south east);
			\node[layernode, anchor=west, minimum width=0.3\wnode, minimum height=3\hnode] (fnn) at ([xshift=2\vseg]s3.east) {\rotatebox{-90}{\small FNN}};
			\node[layernode, anchor=west, minimum height=\wnode, minimum width=0.2\wnode] (hat) at ([xshift=1\vseg]fnn.east) {\tiny$\widehat{z}_j$};
			\draw[decorate] (hat.north east) to node [auto] {\tiny$d$} (hat.south east);
			\foreach \x in {1,2,3}
			{
				\draw[->, thin] (h\x.east) -- (i\x.west);
			}
			\foreach \x in {1,3}
			{
				\draw[->, thin] (i\x.east) -- (p\x1.west);
			}
			\draw[->, thin] (i2.east) to node [auto] {\tiny$\sigma$} (p21.west);
			\draw[-latex', double] (p25.east) to node [auto] {\tiny SoftMax} (a21.west);
			\draw[-latex', double] (w1.south) -- (i3.north);
			\draw[-latex', double] (w2.south) -- (p33.north);
			\foreach \x in {1,2,3}
			{
				\draw[->, thin, densely dotted] (a\x5.east).. controls +(east:1\vseg) and +(west:1\vseg) ..(s5.west);
			}
			\draw[-latex', double] ([xshift=\vseg]s3.east) -- (fnn.west);
			\draw[-latex', double] (fnn.east) -- (hat.west);
			
			\node[anchor=south] () at ([yshift=0\hseg]h3.north) {$1$D};
			\node[anchor=south] () at (s1.north) {$2$D};
			
			\end{scope}
			\end{tikzpicture}\end{center}
		
		\begin{center}
			\vspace{-0.5em}
			\caption{An example of 3-layer representation fusion based on self-attention with 5 hops.}
			\label{fig:compute}
			\vspace{-1.0em}
		\end{center}
	\end{figure*}
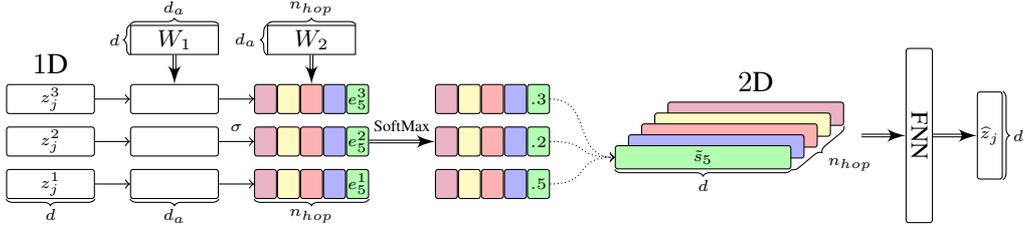

	\textbf{Fusion by Pooling} \hspace{0.4cm} The simplest way of representation fusion is pooling. Here we choose average pooling (avg-pooling) because it shows better results than other pooling methods in our preliminary experiments. Given a sequence of layer outputs at position $j$, the output of average pooling-based fusion is $\phi(Z_j)=\frac{1}{L} \sum_{l=1}^{L} z_j^l$. In general, this way can be seen as the natural generalization of residual connections \cite{he2016identity,he2016deep}. It accumulates all layer features at once when going through the stack, rather than performing identity mapping between adjacent layers at each time. Also, this method is nonparametric and simply assigns an equal weight to each layer in fusion.
	
	\noindent \textbf{Fusion by Feed-forward Neural Network} \hspace{0.4cm} A more sophisticated method is to use feed-forward neural networks (FNNs) for fusion. To this end, we first concatenate all layer outputs to form a big vector. Then, we feed the vector into a single hidden layer FNN with $d_f$ hidden units. Let $d$ be the size of the layer output vector (i.e., $d=|z_j^l|\ \ \forall j, l$). The FNN transforms an $L*d$ dimensional vector to a $d$ dimensional vector. This is good because we do not need to change the size of the downstream components. Moreover, we can learn a better fusion function by the non-linear activation functions used in FNNs. It should be noted that this method can be seen as a special case of dense connection, that is we only concatenate all the features of predecessors in the topmost layer rather than every intermediate layer. As a result, it can reduce the connection complexity from $\mathcal{O}(L^2)$ to $\mathcal{O}(L)$.
	
	\noindent \textbf{Fusion by Multi-hop Self Attention} \hspace{0.4cm} In addition to the pooling and FNN-based methods, we design another fusion function based on the self-attention model with hops \footnote{ In this work, the term "hop" denotes an independent attention computation, which has the same meaning as \newcite{lin2017structured}. More specifically, one hop can generate a vector as the attention distribution over the layers. And different hops may generate the distinguished distributions.} (Figure \ref{fig:compute}). The motivation is straightforward. Different layers reflect different aspects of sentence representation. An ideal way is to take correlations between layers into account, and model the correlation strength (or weight) in generating the fused result. This agrees with the nature of the attention mechanism well \cite{bahdanau2014neural}. Beyond this, self-attention with hops can generate a 2D output that explains different meanings of the input vectors. Previous work has reported that 2D matrix-based representations outperform 1D vector-based counterparts in sentence embedding \cite{lin2017structured}. So it is worth a study on this issue in NMT.
	
	The idea is that we apply self-attention with multiple hops vertically over the layer sequence (along the axis of depth), and extract a part of layer information by each hop. The new problem that arises here is that we need to perceive the temporal order information in sequence \cite{shen2017disan}. Unlike \newcite{vaswani2017attention}'s work, we do not resort to a fixed positional encoding generated by sine and cosine functions of different frequencies. Instead, we learn an embedding of the layer index, as used in positional embedding \cite{gehring2017aconv}. The output of layer embedding is of size $d$, so it is compatible with the layer representation vector $z_j^l$. Also, we use a shared layer embedding for the encoder and the decoder for simple implementation.
	
	More formally, let $E^l \in \mathbb{R}^{d}$ be the embedding of layer $l$. We inject layer embedding into the original representation of the layer, like this:
	
	\begin{equation}
	\tilde{z}^l = z^l + E^l
	\end{equation}
	
	\noindent where $\tilde{z}^l \in \mathbb{R}^{d}$ is the new representation that encodes both the content and index information of layer $l$. Then the self-attention method is performed over $(\tilde{z}^1,...,\tilde{z}^L)$. For hop $p$, the energy of attention weight is computed by an FNN with a single hidden layer,
	
	\begin{equation}
	e = W_{2}(\sigma(W_{1}{\cdot [\tilde{z}^1,...,\tilde{z}^L]}^T))
	\end{equation}
	
	\noindent where $[\tilde{z}^1,...,\tilde{z}^L] \in \mathbb{R}^{d \times L}$ is the 2D layout of layer representations,  $W_{1} \in \mathbb{R}^{d \times d_a}$ and $W_{2} \in \mathbb{R}^{d_a \times n_{hop}}$ are model parameters, $\sigma$ is the activation function, and $d_a$ is the size of inner hidden state. In this model, each hop has an independent parameter vector corresponding to one column of $W_{2}$, whereas  $W_{1}$ is shared across different layers to make efficient use of memory \footnote{$W_1$ can be unique for each layer. We discuss this issue in Section \ref{sec:zh2en}.}. After that, we obtain the attention weight vector $\textrm{a}_p=(a_p^1,...,a_p^L)$ of hop $p$ by Softmax. For each layer index $l$, we have
	
	\begin{equation}
	\label{eq:attention-softmax}
	a_p^l = \frac{\exp(e_p^l)}{\sum_{l'=1}^{L} \exp(e_p^{l'})}
	\end{equation}
	
	The self-attention model generates an intermediate representation by dot-production ($\odot$) of the weights and the layer representations.
	
	\begin{equation}
	\label{eq:attention-dot}
	\tilde{s}_p = a_p \odot [\tilde{z}^1,...,\tilde{z}^L]
	\end{equation}
	
	In this way, all the hops compose a 2D matrix $\tilde{s}= [\tilde{s}_1,...,\tilde{s}_{n_{hop}}] \in \mathbb{R}^{d \times n_{hop}}$. This matrix is fed into another single hidden layer FNN with $d_f$ hidden units for the final output. It is worth noting that the model presented here can be enhanced by using an independent weight $a_{p;k}^l$ for each dimension $k$ of $\tilde{z}^l$ (call it feature-wise fusion). But this method is computationally expensive. In our experiments we do not observe improvements by this method. Therefore, we choose the version described in Eqs. (\ref{eq:attention-softmax}-\ref{eq:attention-dot}) for the empirical study.
	
	Table \ref{table:methods} shows a comparison of the methods used in this work. For good convergence, we use layer normalization after the fusion layer in this work.
	
	\begin{table*}[t]
		\begin{center}
			\begin{tabular}{r | l | l | l | l }
				\hline
				Entry & Fusion Function & Low-level   & Parametric & 2D \\
				&      & accessible  &            & represent. \\ \hline
				Baseline & $\phi(Z_j)=z_j^L$ & No & No & No \\
				Avg-pooling & $\phi(Z_j)=\frac{1}{L} \sum_{l=1}^{L} z_j^l$ & Yes & No & No \\
				FNN-based & $\phi(Z_j)=\textrm{FNN}([z_j^1,...,z_j^L])$ & Yes & Yes & No \\
				Self-att.-based & $\phi(Z_j)=\textrm{FNN}(\textrm{SelfAtt}([z_j^1,...,z_j^L]))$ & Yes & Yes & Yes \\
				\hline
			\end{tabular}
		\end{center}
		\vspace{-0.5em}
		\caption{Comparison of the fusion functions used in this work.}
		\label{table:methods}
		\vspace{-0.5em}
	\end{table*}

	\section{Experiments}

	\noindent We evaluate our proposed approach on German-English and Chinese-English translation tasks. 
	\subsection{Setup}
	\noindent For German-English translation, we use the data of the IWSLT 2014 German-English track \cite{cettolo2014report}. We follow \newcite{ranzato2015sequence}'s work for preprocessing. We use a joint source and target byte-pair encoding with 10k merge operations \cite{sennrich2015neural}. The source and target vocabulary sizes are 8,389 and 6,428 respectively. We remove the sentences with more than 175 words or 100 sub-word units. This results in 160K sentence pairs for training. We randomly sample 7K sentences from the training data for held-out validation, and concatenate dev2010, dev2012, tst2010, tst2011, and tst2012 for test.
	
	For Chinese-English translation, we use parts of the bitext provided within NIST12 OpenMT\footnote{LDC2000T46, LDC2000T47, LDC2000T50, LDC2003E14, LDC2005T10, LDC2002E18, LDC2007T09, LDC2004T08}. We choose NIST 2006 (MT06) as the validation set, and 2004 (MT04), 2005 (MT05), 2008 (MT08) as the test sets. All Chinese sentences are word segmented using the tool provided within NiuTrans \cite{xiao2012niutrans}. All sentences of more than 50 words are removed. The resulting training corpus consists of 1.85M sentence pairs, with 39.42M Chinese words and 44.92M English words on each language side. We limit the vocabularies to the most frequent 30K words, covering approximately 98.16\% and 99.17\% of the Chinese and English words respectively. All out-of-vocabulary words are replaced with $<$UNK$>$. We report results without any UNK-replacement techniques \cite{luong2015address}.
	
	\subsection{Implementation Details}
	\begin{table*}[t]
		\begin{center}
			\renewcommand\arraystretch{0.9}
			\begin{tabular}{ll | c | c | c }
				\hline
				\multicolumn{2}{c |}{\textbf{System}} & \multicolumn{1}{c |}{\textbf{\#Param.}} &
				\multicolumn{1}{c |}{\textbf{Valid.}} & \multicolumn{1}{c }{\textbf{Test}} \\
				\hline 
				
				\multicolumn{1}{l}{\multirow{7}{*}{Existing Systems}} &
				\multicolumn{1}{|l|}{\footnotesize{RNN-MIXER}  \cite{ranzato2015sequence}} 		& - 	& - 	& 20.73 \\
				~ & \multicolumn{1}{|l|}{\footnotesize{RNN-BSO \cite{wiseman2016sequence}}}  			& - 	& - 	& 26.36 \\
				~ & \multicolumn{1}{|l|}{\footnotesize{RNN-AC\cite{bahdanau+al-2016-actorcritic}}}  				& - 	& - 	& 28.53 \\
				~ & \multicolumn{1}{|l|}{\footnotesize{RNN-NPMT \cite{huang2017neural}}}  				& - 	& - 	& 28.96 \\
				~ & \multicolumn{1}{|l|}{\footnotesize{RNN-NPMT + LM \cite{huang2017neural}}}  		& - 	& - 	& 29.16 \\
				\cdashline{2-5}
				~ & \multicolumn{1}{|l|}{\footnotesize{ConvSeq2Seq-MLE \cite{edunov2017classical}}}  	& - 	& 32.96 & 31.74 \\
				~ & \multicolumn{1}{|l|}{\footnotesize{ConvSeq2Seq-Risk \cite{edunov2017classical}}}  	& - 	& 33.91 & 32.85 \\
				\hline
				
				\multicolumn{1}{l}{\multirow{4}{*}{Baselines}} &
				\multicolumn{1}{|l|}{\footnotesize{Transformer-MLE}}  								& 10.97M & 33.58 & 31.75 \\
				~ & \multicolumn{1}{|l|}{\footnotesize{\hspace{0.3cm} +RestartAdam}}  					& 10.97M & 34.14 & 32.67 \\
				~ & \multicolumn{1}{|l|}{\footnotesize{\hspace{0.3cm} +RestartAdam-4Layers}}  			& 12.82M & 34.20 & 32.57 \\
				~ & \multicolumn{1}{|l|}{\footnotesize{\hspace{0.3cm} +RestartAdam-6Layers}}  			& 16.50M & 34.35 & 32.97 \\
				\hline
				
				\multicolumn{1}{l}{\multirow{11}{*}{MLRF Systems}} &
				\multicolumn{1}{|l|}{\footnotesize{Enc-AVG}}  										& 10.97M & 34.34 & 32.71 \\
				~ & \multicolumn{1}{|l|}{\footnotesize{Enc-FNN} } 										& 11.63M & 34.55 & 33.31 \\
				~ & \multicolumn{1}{|l|}{\footnotesize{Enc-SA ($n_{hop}$=4)}}  						& 11.90M & 34.48 & 33.06 \\
				~ & \multicolumn{1}{|l|}{\footnotesize{Enc-SA ($n_{hop}$=6)}}  						& 12.16M & 34.61 & 33.40 \\
				
				\cdashline{2-5}
				~ & \multicolumn{1}{|l|}{\footnotesize{Dec-AVG} } 										& 10.97M & 34.02 & 32.80 \\
				~ & \multicolumn{1}{|l|}{\footnotesize{Dec-FNN} } 										& 11.63M & 34.38 & 32.93 \\
				~ & \multicolumn{1}{|l|}{\footnotesize{Dec-SA ($n_{hop}$=4)}}  						& 11.90M & \textbf{34.83} & 33.29 \\
				~ & \multicolumn{1}{|l|}{\footnotesize{Dec-SA ($n_{hop}$=6)}}  						& 13.47M & \textbf{34.73} & 33.54 \\
				
				\cdashline{2-5}
				~ & \multicolumn{1}{|l|}{\footnotesize{Both-FNN}}  									& 12.29M & 34.30 & 32.66 \\
				~ & \multicolumn{1}{|l|}{\footnotesize{Both-SA ($n_{hop}=4$)} } 										& 12.82M & 34.59 & 33.46 \\
				~ & \multicolumn{1}{|l|}{\footnotesize{Both-FNN-SA ($n_{hop}$=4)}}  					& 12.55M & 34.55 & \textbf{33.59} \\
				\hline
			\end{tabular}
			
			\vspace{-0.0em}
			\caption{BLEU scores [\%] on IWSLT German-English translation.}
			\label{table:de2en-result}
			\vspace{-0.5em}
		\end{center}
	\end{table*}
	
	For German-English systems, the model consists of a 3-layer encoder and a 3-layer decoder with $d=256$ and 1024 hidden units in the FNN sub-layer. For our approach, we set $d_a=1024$ and $d_f=512$. %
	Dropout (rate$=0.1$) is used for regularization.
	We initialize all word/sub-word embedding matrices using a normal distribution $\mathcal N(0, d^{-0.5})$, while initialize the layer embedding matrix using a uniform distribution $\mathcal U(-0.1, 0.1)$. The weight and bias in layer-normalization are initialized to constants 1 and 0. All other parameters are initialized from $\mathcal U(-\frac{1}{\sqrt{fan_{in}}}, \frac{1}{\sqrt{fan_{in}}})$, where $fan_{in}$ is the input size of the parameter matrix.
	We use negative Maximum Likelihood Estimation (MLE) as loss function, and train all the models using Adam \cite{kingma2014adam} with $\beta_1=0.9$, $\beta_2=0.98$, and $\epsilon=10^{-9}$. We run training for 40 epochs with a mini-batch of 80. The learning rate is scheduled as described in \cite{vaswani2017attention}:
	$ lr=d^{-0.5} \cdot min(t^{-0.5}, t \cdot \textrm{16k}^{-1.5}), $
	where $t$ is the step number. After that, we restart Adam and continue the training for additional 20 epochs with a fixed learning rate $5e^{-5}$ and a smaller mini-batch of 32 \cite{denkowski2017stronger}.
	At test time, translations are generated by beam search with length normalization. By tuning on the validation set, we use a beam of width 8 and a length normalization weight of 1.6.
	
	For Chinese-English systems, we use a 6-layer encoder and a 6-layer decoder, with $d=512$ and 2048 hidden units in the FNN sub-layer. We restart Adam after 10 epochs and train the model for 5 additional epochs. A beam of width 12 and a length normalization weight of 1.3 are employed.
	
	 Note that the network equipped with our fusion layer increases a fraction of the computation cost than original one. E.g., in Chinese-English translation, the training speed reduces from 2.6 batches/second (baseline) to 2.4 batches/second (Dec-SA, ref. row 11 in Table \ref{table:zh2en-result}).
	
	\subsection{Results on German-English Translation}
	\noindent Table \ref{table:de2en-result} shows the BLEU scores of various NMT systems\footnote{All BLEU results are case-insensitive. For a fair comparison with previous work, we run \textit{multi-bleu.perl} for German-English translation, and run \textit{mteval-v13a.pl} for Chinese-English translation.}. For comparison, we list previous results on the same data set. First of all, our baseline is good. The base setting of our baseline system can lead to a similar score to ConvSeq2Seq which is trained for more epochs \cite{edunov2017classical}. The baseline is stronger when Adam restart is adopted. It outperforms most of previous systems on this task (only lower than the best reported system by 0.2 BLEU points).
	
	Then, we test different fusion functions (AVG, FNN and SA) on the encoder side (Enc), the decoder side (Dec) and both of them (Both). We see that avg-pooling does not yield promising improvements due to the relatively low expressive power. In addition, the representation fusion on the decoder side is more effective than that on the encoder side.
	A possible reason is that the prediction can benefit more from the lower-level layers in the decoder due to the ``shorter" distance from fusion to prediction. This agrees with the result that representation fusion on both sides (Both-SA) does not improve Enc-SA and Dec-SA significantly. The best result is achieved when we use FNN-based fusion on the encoder side and self-attention-based fusion on the decoder side (Both-FNN-SA). It outperforms the 3-layer baseline by 0.92 BLEU points.
	Also, we increase the number of layers to 4 and 6 to examine whether the improvement comes from the additional model parameters introduced by the added fusion layer. It results in stronger baselines, but they still underperform the Both-FNN-SA system which has fewer parameters.
	
	\subsection{Results on Chinese-English Translation}
	\label{sec:zh2en}
	\begin{table*}[tbp]
		\begin{center}
			\renewcommand\arraystretch{0.9}
			\begin{tabular}{ll | c | c c c c}
				\hline
				\multicolumn{2}{c |}{} & \multicolumn{1}{c |}{\textbf{Valid.}} &
				\multicolumn{4}{c }{\textbf{Test}} \\
				\multicolumn{2}{c |}{\textbf{System}}  	& \textbf{\footnotesize{MT06}} & \textbf{\footnotesize{MT04}} & \textbf{\footnotesize{MT05}} & \textbf{\footnotesize{MT08}}  & \textbf{\footnotesize{Ave.}} \\
				\hline 
				\multicolumn{1}{l}{\multirow{3}{*}{Open Source Systems}} &
				\multicolumn{1}{|l| }{\footnotesize{Nematus-4Layers}}  & 40.49 & 46.83 & 39.40 & 32.73 & 39.65 \\
				~ & \multicolumn{1}{|l| }{\footnotesize{NMTTutorial-GNMT}} & 41.29 & 47.43 & 40.18 & 33.67 & 40.43 \\
				~ & \multicolumn{1}{|l| }{\footnotesize{T2T}}		& 41.87	& 47.42	& 41.09	& 34.39 & 40.97 \\
				
				\hline
				\multicolumn{1}{l}{\multirow{2}{*}{Baselines}} &
				\multicolumn{1}{|l| }{\footnotesize{Transformer-MLE}} & 40.60 & 47.70 & 40.33 & 34.70 & 40.91 \\
				~ &  \multicolumn{1}{|l| }{\footnotesize{\hspace{0.3cm} +RestartAdam}}  	& 41.82 & 48.09 & 40.76 & 34.75 & 41.20 \\
				\hline
				
				\multicolumn{1}{l}{\multirow{8}{*}{MLRF Systems}} &
				\multicolumn{1}{|l| }{\footnotesize{Enc-AVG}} 	& 42.17 & 48.35	& 41.24	& 34.93 & 41.51 \\
				~ & \multicolumn{1}{|l| }{\footnotesize{Enc-FNN}} & 40.81	& 47.44	& 40.03	& 33.55	& 40.34 \\
				~ & \multicolumn{1}{|l| }{\footnotesize{Enc-SA ($n_{hop}=4$)}} & 40.17 & 46.79 & 39.55 & 32.80 & 39.71 \\
				\cdashline{2-7}
				~ & \multicolumn{1}{|l| }{\footnotesize{Dec-AVG}}  & 41.43 & 48.38	& 41.19	& 34.66	& 41.49 \\
				~ & \multicolumn{1}{|l| }{\footnotesize{Dec-FNN}}  & 41.97 & 48.14 & 41.36 & 34.38 & 41.23 \\
				~ & \multicolumn{1}{|l| }{\footnotesize{Dec-SA ($n_{hop}=4$)}} & 42.13 & 48.48 & \textbf{41.51} & 34.79 & 41.59 \\
				~ & \multicolumn{1}{|l| }{ \footnotesize{\hspace{0.3cm} + indep. $W_{1}$}} & \textbf{42.26} & \textbf{48.75} & 41.44 & \textbf{35.08} & \textbf{41.76}  \\
				
				\cdashline{2-7}
				~ & \multicolumn{1}{|l| }{ \footnotesize{Both-AVG-SA ($n_{hop}=4$)}}  & 41.61 & 47.65 & 40.69 & 34.52 & 40.95 \\
				\hline

			\end{tabular}
			
			\vspace{-0.0em}
			\caption{BLEU scores [\%] on NIST Chinese-English translation.}
			\label{table:zh2en-result}
			\vspace{-0.5em}
		\end{center}
	\end{table*}
	For Chinese-English translation, we compare our systems with three open source systems: Nematus \cite{sennrich-EtAl:2017:EACLDemo}, NMTTutorial \cite{luong17} and T2T \footnote{\url{https://github.com/tensorflow/tensor2tensor}}. The first two are based on RNN, while the last one is based on Transformer. Table \ref{table:zh2en-result} shows that our baseline with restarting Adam is a bit better than T2T. Like in German-English translation, fusion on the decoder side is superior than that on the encoder side. But both Enc-FNN and Enc-SA are worse than the baseline. We suspect that adding more layers with non-linear transformations makes it more difficult to train the deep network (6 layers in Chinese-English translation) than the shallow counterpart (3 layers in German-English translation). This problem is more evident here because the Chinese-English systems have many more parameters than the German-English systems and are more difficult to optimize. Interestingly, we see that the simple avg-pooling fusion method works well in both encoder and decoder. In addition, using independent $W_1$ (in self-attention) for each layer is helpful for better fitting. It indicates that the MLRF method can benefit from more sophisticated design of the self-attention model.
	
	\subsection{Effect on Attention Hops}
		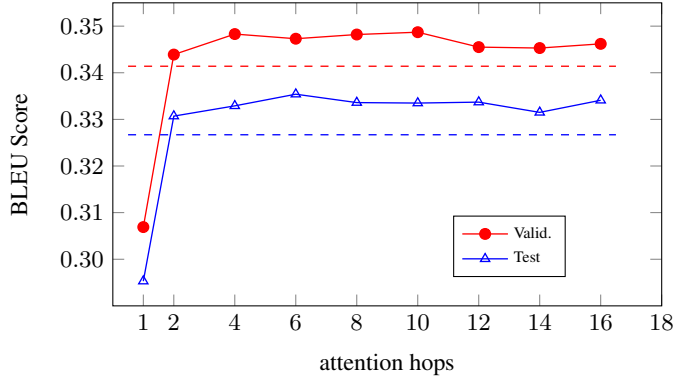
\begin{figure*}[tbp]
		\centering
		
		\begin{tikzpicture}{baseline}
		\footnotesize{
			\begin{axis}[
			width=.55\textwidth,
			height=.35\textwidth,
			legend style={at={(0.62,0.2)}, anchor=south west},
			xlabel={\footnotesize{attention hops}},
			ylabel={\footnotesize{BLEU Score}},
			ylabel style={yshift=-0em},xlabel style={yshift=0.0em},
			yticklabel style={/pgf/number format/precision=2,/pgf/number format/fixed zerofill},
			ymin=0.29,ymax=0.355, ytick={0.30, 0.31, 0.32, 0.33, 0.34, 0.35},
			xmin=0,xmax=18,xtick={1,2,4,6,8,10,12,14, 16, 18},
			legend style={yshift=-12pt, legend plot pos=left,font=\tiny,cells={anchor=west}}
			]
			
			\addplot[red,mark=otimes*,line width=0.5pt] coordinates {(1,0.3069) (2,0.3439) (4,0.3483) (6,0.3473) (8,0.3482) (10,0.3487) (12,0.3455) (14,0.3453) (16,0.3462)};
			\addlegendentry{Valid.}
			\addplot[blue,mark=triangle,line width=0.5pt] coordinates {(1,0.2953) (2,0.3307) (4,0.3329) (6,0.3354) (8,0.3336) (10,0.3335) (12,0.3337) (14,0.3315) (16,0.3341)};
			\addlegendentry{Test}
			
			\addplot[dashed,domain=0.5:16.5, red, line width=0.5pt]{0.3414};
			
			\addplot[dashed,domain=0.5:16.5, blue, line width=0.5pt]{0.3267};
			
			\end{axis}
		}
		\end{tikzpicture}
		\caption{BLEU scores under various attention hops in Dec-SA on German-English translation. Dashed lines denotes the baseline. }
		\label{fig:hops}
	\end{figure*}
	Figure \ref{fig:hops} shows the BLEU curves by varying $n_{hop}$ on German-English translation with the Dec-SA approach. Clearly, increasing the number of hops improves the system. The improvement is significant when $n_{hop} < 6$ thanks to the 2D representations extracted by multiple hops. Note that when $n_{hop}=1$, the 2D sentence representation is degraded into the 1D representation. It results in a dramatic decrease in BLEU. However, larger $n_{hop}$ does not make further improvements due to the redundant information in hops and potential overlaps between them.
	
	\subsection{Importance of Word Embedding and Layer Embedding}

	\begin{table*}[tbp]
		\begin{center}
			\renewcommand\arraystretch{0.85}
			\begin{tabular}{l | c c | c }
				\hline
				\multicolumn{1}{c |}{\textbf{System}} & \multicolumn{1}{c }{\textbf{Word Emb.}} &
				\multicolumn{1}{c |}{\textbf{Layer Emb.}} &
				\multicolumn{1}{c }{\textbf{Test}} \\
				\hline 
				\footnotesize{Enc-FNN} 			& $\surd$	& $\surd$	& 33.31 \\
				& $\surd$	& 			& 33.17 \\
				& 			& $\surd$	& $33.07^*$ \\
				\hline
				\footnotesize{Dec-SA} 			& $\surd$	& $\surd$	& 33.29 \\
				& $\surd$	&		 	& $33.06^*$ \\
				& 			& $\surd$	& 33.21 \\
				
				\hline
				\footnotesize{Both-FNN-SA} 		& $\surd$	& $\surd$	& 33.59 \\
				& $\surd$	&		 	& 33.55 \\
				& 			& $\surd$	& $33.16^*$ \\
				
				\hline
			\end{tabular}
			
			\vspace{-0.0em}
			\caption{BLEU scores on German-English translation when fusing with (denoted by $\surd$)/without word embedding and layer embedding in encoder, decoder, and both of them. * denotes significant performance reduction ($>$ 0.2 BLEU) than baseline (with both word embedding and layer embedding).}
			\label{table:embedding}
			\vspace{-0.0em}
		\end{center}
	\end{table*}
	
	Then, we study the model behavior with/without word embedding or layer embedding on German-English translation. Table \ref{table:embedding} shows that removing either word embedding or layer embedding harms the Enc-FNN system. The impact of word embedding is a bit more than that of layer embedding. As a contrast, Dec-SA is more sensitive to the remove of layer embedding. We attribute it to the unawareness of the temporal order information in the self-attention mechanism. More interestingly, when we fuse representations on both sides, almost no BLEU reduction is observed in Both-FNN-SA. It is true even if layer embedding is removed. This result indicates that the layer indexing information can be delivered from the FNN fusion layer in the encoder to the decoder in an implicit way. Also, a significant improvement can be achieved by fusing with the original word embedding features. It agrees with the result reported in \cite{xiong2017multi}.
	
	\subsection{Regularization}

		\begin{figure*}[tbp]
			\centering
			\begin{tikzpicture}{baseline}
			\footnotesize{
				\begin{axis}[
				width=.55\textwidth,
				height=.35\textwidth,
				xlabel={\footnotesize{update steps}},ylabel={\footnotesize{Acc. (\%)}},
				ylabel style={yshift=0em},xlabel style={yshift=0em},
				yticklabel style={/pgf/number format/precision=1,/pgf/number format/fixed zerofill},
				ymin=55,ymax=74.5,
				legend style={yshift=0.1pt,legend plot pos=left,legend pos=south east, font=\tiny, cells={anchor=west}, legend style={column sep=1pt, inner sep=0pt}}
				]
				
				\addplot[red] table {de2en.baseline.acc};
				\addlegendentry{baseline(p=0.1)}
				\addplot[blue] table {de2en.baseline-dropout2.acc};
				\addlegendentry{p=0.2}C
				\addplot[ugreen] table {de2en.decoder-selfatt.acc};
				\addlegendentry{Dec-SA}
				
				\addplot[red,dashed] table {de2en.baseline.train};
				\addplot[blue,dashed] table {de2en.baseline-dropout2.train};
				\addplot[ugreen,dashed] table {de2en.decoder-selfatt.train};
				
				\end{axis}
			}
			\end{tikzpicture}
			
			\caption{Curve of Accuracy in training-set (dashed lines) and validation-set (solid lines) along with update steps. $p$ is dropout rate. }
			\label{fig:regularization}
	\end{figure*}
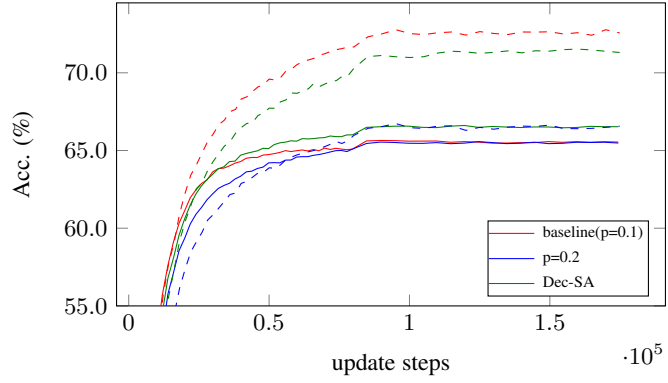
	

	Also, we plot the curves of accuracy on the training and validation sets in Figure \ref{fig:regularization} (German-English translation). It is obvious that the baseline with dropout rate $p=0.1$ (red) has a significant lower BLEU score when we switch from training to validation, implying that the model somehow over-fits the data. When we use a larger $p=0.2$ (blue) for dropout, the accuracy on the training set decreases dramatically, but no promising improvement is observed on the validation set. On the contrary, Dec-SA seems to have a better regularization effect. Although Dec-SA has lower accuracy in training than the baseline, it outperforms the baseline on the validation set. It might be due to the introduced direct connections from lower-level layers to the top of the stack. These connections make the model easier to fit the data because they have no dependence on the non-linear transformations in the stacked layers.

	\subsection{Layer Attention Visibility}
	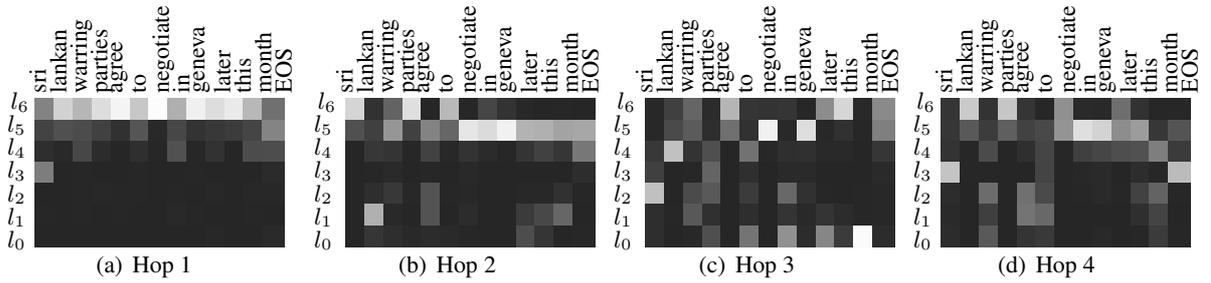
\begin{figure*}[t]
		\begin{center}
			\renewcommand{\arraystretch}{0}
			\setlength{\tabcolsep}{0.5mm}
			\setlength{\fboxsep}{1.4mm} 
			
			\begin{tabular}{C{.24\textwidth}C{.24\textwidth}C{.24\textwidth}C{.24\textwidth}}
				\setlength{\tabcolsep}{0pt}

				\subfigure [\footnotesize{Hop 1}] {
					
					\begin{tabular}{cc}
						\setlength{\tabcolsep}{0pt}
						~
						&
						
						\begin{tikzpicture}
						\begin{scope}						
						\node [inner sep=0pt] (w1) at (0,0) {\rotatebox{90}{\footnotesize{sri}} };
						
						\foreach \x/\y/\z in {2/1/lankan, 3/2/warring, 4/3/parties, 5/4/agree, 6/5/to, 7/6/negotiate, 8/7/in, 9/8/geneva, 10/9/later, 11/10/this, 12/11/month, 13/12/EOS}
						{
							\node [inner sep=0pt,anchor=south west] (w\x) at ([xshift=0.7em]w\y.south west) {\rotatebox{90}{\footnotesize{\z}} };
						} 
						\end{scope}
						\end{tikzpicture}
						
						\\
						
						\renewcommand\arraystretch{0.6}
						\begin{tabular}{c}
							\setlength{\tabcolsep}{0pt}
							\footnotesize{$l_6$} \\
							\footnotesize{$l_5$} \\
							\footnotesize{$l_4$} \\
							\footnotesize{$l_3$} \\
							\footnotesize{$l_2$} \\
							\footnotesize{$l_1$} \\
							\footnotesize{$l_0$} \\
						\end{tabular}
						&
						\begin{tabular}{*{13}{Q}}
							0.4301 & 0.7933 & 0.6630 & 0.8416 & 0.9468 & 0.7395 & 0.9841 & 0.6305 & 0.9285 & 0.8433 & 0.9012 & 0.6703 & 0.3388 \\
							
							0.1312 & 0.1918 & 0.1758 & 0.1234 & 0.0488 & 0.2162 & 0.0132 & 0.1534 & 0.0593 & 0.0837 & 0.0984 & 0.1430 & 0.4319 \\
							
							0.0355 & 0.0142 & 0.1502 & 0.0249 & 0.0001 & 0.0407 & 0.0021 & 0.1976 & 0.0023 & 0.0530 & 0.0003 & 0.1752 & 0.1783 \\
							
							0.3994 & 0.0001 & 0.0031 & 0.0009 & 0.0004 & 0.0014 & 0.0003 & 0.0008 & 0.0052 & 0.0000 & 0.0000 & 0.0003 & 0.0209 \\
							
							0.0003 & 0.0006 & 0.0045 & 0.0078 & 0.0020 & 0.0013 & 0.0000 & 0.0046 & 0.0005 & 0.0102 & 0.0000 & 0.0095 & 0.0100 \\
							
							0.0035 & 0.0000 & 0.0036 & 0.0010 & 0.0019 & 0.0001 & 0.0003 & 0.0131 & 0.0031 & 0.0003 & 0.0000 & 0.0017 & 0.0011 \\
							
							0.0000 & 0.0000 & 0.0000 & 0.0004 & 0.0000 & 0.0008 & 0.0000 & 0.0000 & 0.0011 & 0.0095 & 0.0001 & 0.0000 & 0.0190 \\

						\end{tabular} 
						
					\end{tabular}
				}
				&
				
				\subfigure [\footnotesize{Hop 2}] {
					\setlength{\tabcolsep}{0pt}
					\begin{tabular}{cc}
						\setlength{\tabcolsep}{0pt}
						~
						&
						
						\begin{tikzpicture}
						\begin{scope}						
						\node [inner sep=0pt] (w1) at (0,0) {\rotatebox{90}{\footnotesize{sri}} };
						
						\foreach \x/\y/\z in {2/1/lankan, 3/2/warring, 4/3/parties, 5/4/agree, 6/5/to, 7/6/negotiate, 8/7/in, 9/8/geneva, 10/9/later, 11/10/this, 12/11/month, 13/12/EOS}
						{
							\node [inner sep=0pt,anchor=south west] (w\x) at ([xshift=0.7em]w\y.south west) {\rotatebox{90}{\footnotesize{\z}} };
						} 
						\end{scope}
						\end{tikzpicture}
						
						\\
						
						\renewcommand\arraystretch{0.6}
						\begin{tabular}{c}
							\setlength{\tabcolsep}{0pt}
							\footnotesize{$l_6$} \\
							\footnotesize{$l_5$} \\
							\footnotesize{$l_4$} \\
							\footnotesize{$l_3$} \\
							\footnotesize{$l_2$} \\
							\footnotesize{$l_1$} \\
							\footnotesize{$l_0$} \\
						\end{tabular}
						&
						
						\setlength{\tabcolsep}{0pt}
						\begin{tabular}{*{13}{Q}}
							0.8069 & 0.0539 & 0.2805 & 0.8525 & 0.0004 & 0.6779 & 0.0046 & 0.1472 & 0.0579 & 0.0100 & 0.0005 & 0.0027 & 0.0006 \\
							
							0.1917 & 0.1192 & 0.5153 & 0.1225 & 0.4299 & 0.2934 & 0.8819 & 0.8317 & 0.9364 & 0.6483 & 0.6380 & 0.5782 & 0.5935 \\
							
							0.0007 & 0.0794 & 0.0615 & 0.0033 & 0.1287 & 0.0062 & 0.0602 & 0.0197 & 0.0052 & 0.0613 & 0.1256 & 0.0740 & 0.3715 \\
							
							0.0007 & 0.0032 & 0.0157 & 0.0006 & 0.0012 & 0.0006 & 0.0113 & 0.0004 & 0.0000 & 0.0000 & 0.0006 & 0.0009 & 0.0308 \\
							
							0.0001 & 0.0340 & 0.0796 & 0.0074 & 0.2088 & 0.0029 & 0.0099 & 0.0009 & 0.0004 & 0.0064 & 0.0101 & 0.0505 & 0.0018 \\
							
							0.0000 & 0.6374 & 0.0129 & 0.0003 & 0.2048 & 0.0001 & 0.0278 & 0.0000 & 0.0000 & 0.0941 & 0.1576 & 0.2935 & 0.0004 \\
							
							0.0000 & 0.0729 & 0.0345 & 0.0133 & 0.0263 & 0.0188 & 0.0043 & 0.0001 & 0.0000 & 0.1799 & 0.0675 & 0.0001 & 0.0014 \\
						\end{tabular}

					\end{tabular}
					
				}
				
				&
				
				\subfigure [\footnotesize{Hop 3}] {
					\setlength{\tabcolsep}{0pt}
					\begin{tabular}{cc}
						~
						&
						\begin{tikzpicture}
						\begin{scope}						
						\node [inner sep=0pt] (w1) at (0,0) {\rotatebox{90}{\footnotesize{sri}} };
						
						\foreach \x/\y/\z in {2/1/lankan, 3/2/warring, 4/3/parties, 5/4/agree, 6/5/to, 7/6/negotiate, 8/7/in, 9/8/geneva, 10/9/later, 11/10/this, 12/11/month, 13/12/EOS}
						{
							\node [inner sep=0pt,anchor=south west] (w\x) at ([xshift=0.7em]w\y.south west) {\rotatebox{90}{\footnotesize{\z}} };
						} 
						\end{scope}
						\end{tikzpicture}
						
						\\
						
						\renewcommand\arraystretch{0.6}
						\begin{tabular}{c}
							\setlength{\tabcolsep}{0pt}
							\footnotesize{$l_6$} \\
							\footnotesize{$l_5$} \\
							\footnotesize{$l_4$} \\
							\footnotesize{$l_3$} \\
							\footnotesize{$l_2$} \\
							\footnotesize{$l_1$} \\
							\footnotesize{$l_0$} \\
						\end{tabular}
						&
						\setlength{\tabcolsep}{0pt}
						\begin{tabular}{*{13}{Q}}

							0.0093 & 0.1187 & 0.2911 & 0.0228 & 0.6598 & 0.0702 & 0.0575 & 0.0694 & 0.0032 & 0.4705 & 0.8112 & 0.0137 & 0.4623 \\
							
							0.0099 & 0.1634 & 0.2447 & 0.0375 & 0.3395 & 0.0483 & 0.9330 & 0.0136 & 0.8455 & 0.0073 & 0.0501 & 0.0025 & 0.4085 \\
							
							0.0816 & 0.7134 & 0.0571 & 0.1858 & 0.0000 & 0.3515 & 0.0087 & 0.0382 & 0.0432 & 0.0120 & 0.0123 & 0.0003 & 0.0746 \\
							
							0.1210 & 0.0000 & 0.0026 & 0.2868 & 0.0000 & 0.0414 & 0.0004 & 0.0166 & 0.0193 & 0.0000 & 0.0006 & 0.0001 & 0.0165 \\
							
							0.7079 & 0.0006 & 0.1700 & 0.2343 & 0.0004 & 0.0949 & 0.0003 & 0.3012 & 0.0507 & 0.0055 & 0.0007 & 0.0032 & 0.0079 \\
							
							0.0351 & 0.0038 & 0.2281 & 0.0447 & 0.0002 & 0.0349 & 0.0001 & 0.0655 & 0.0126 & 0.0615 & 0.0287 & 0.0017 & 0.0058 \\
							
							0.0351 & 0.0000 & 0.0065 & 0.1880 & 0.0000 & 0.3587 & 0.0000 & 0.4955 & 0.0253 & 0.4431 & 0.0963 & 0.9786 & 0.0245 \\
						\end{tabular}
						
					\end{tabular}
					
				}
				&
				\subfigure [\footnotesize{Hop 4}] {
					\setlength{\tabcolsep}{0pt}
					
					\begin{tabular}{cc}
						\setlength{\tabcolsep}{0pt}
						~
						&
						
						\begin{tikzpicture}
						\begin{scope}						
						\node [inner sep=0pt] (w1) at (0,0) {\rotatebox{90}{\footnotesize{sri}} };
						
						\foreach \x/\y/\z in {2/1/lankan, 3/2/warring, 4/3/parties, 5/4/agree, 6/5/to, 7/6/negotiate, 8/7/in, 9/8/geneva, 10/9/later, 11/10/this, 12/11/month, 13/12/EOS}
						{
							\node [inner sep=0pt,anchor=south west] (w\x) at ([xshift=0.7em]w\y.south west) {\rotatebox{90}{\footnotesize{\z}} };
						} 
						\end{scope}
						\end{tikzpicture}
						
						\\
						
						\renewcommand\arraystretch{0.6}
						\begin{tabular}{c}
							\setlength{\tabcolsep}{0pt}
							\footnotesize{$l_6$} \\
							\footnotesize{$l_5$} \\
							\footnotesize{$l_4$} \\
							\footnotesize{$l_3$} \\
							\footnotesize{$l_2$} \\
							\footnotesize{$l_1$} \\
							\footnotesize{$l_0$} \\
						\end{tabular}
						&
						\begin{tabular}{*{13}{Q}}
							0.0682 & 0.7567 & 0.0584 & 0.7360 & 0.0682 & 0.0030 & 0.4899 & 0.0168 & 0.0126 & 0.3320 & 0.0548 & 0.0025 & 0.0048 \\
							
							0.0727 & 0.2351 & 0.0993 & 0.2630 & 0.1299 & 0.1616 & 0.5076 & 0.8523 & 0.7954 & 0.4764 & 0.5421 & 0.0782 & 0.2100 \\
							
							0.0572 & 0.0047 & 0.1716 & 0.0008 & 0.0391 & 0.1402 & 0.0013 & 0.1201 & 0.1462 & 0.1854 & 0.1988 & 0.4051 & 0.0948 \\
							
							0.7289 & 0.0021 & 0.0020 & 0.0001 & 0.0005 & 0.1383 & 0.0003 & 0.0073 & 0.0160 & 0.0003 & 0.0321 & 0.0020 & 0.6853 \\
							
							0.0303 & 0.0009 & 0.3205 & 0.0002 & 0.3407 & 0.1670 & 0.0005 & 0.0031 & 0.0156 & 0.0052 & 0.1297 & 0.2813 & 0.0024 \\
							
							0.0155 & 0.0000 & 0.1141 & 0.0000 & 0.3578 & 0.3056 & 0.0000 & 0.0003 & 0.0001 & 0.0000 & 0.0414 & 0.1373 & 0.0006 \\
							
							0.0272 & 0.0005 & 0.2340 & 0.0000 & 0.0639 & 0.0842 & 0.0004 & 0.0001 & 0.0140 & 0.0006 & 0.0012 & 0.0937 & 0.0020 \\

						\end{tabular} 
						
					\end{tabular}
				}

			\end{tabular}
		\end{center}
		
		\begin{center}
			\vspace{-1.5em}
			\caption{A visualization example of Dec-SA with 4 hops in Chinese-English translation. Colors change between white (probability of 1) to black (probability of 0). The source sentence (pinyin) is \textit{\pinyin{Si1li3lan2ka3} \pinyin{jiao1zhan4} \pinyin{shuang1fang1} \pinyin{tong2yi4} \pinyin{ben3yue4} \pinyin{xia4xun2} \pinyin{zai4} \pinyin{ri4nei4wa3} \pinyin{tan2pan4}}.}
			\label{figure:multihop}
			\vspace{-1.3em}
		\end{center}
	\end{figure*}
	A visualization example of layer attention with multiple hops is presented in Figure \ref{figure:multihop}. The attention result is generated from an example of Chinese-English translation by Dec-SA with 4 hops. The x-axis is the target word sequence, and the y-axis is the indexing of layers (including the word embedding layer denoted by $l_0$). We can see that most attention weights focus on the high-level layers like hops 1 and 2, while hops 3 and 4 have more dispersive attentions over different layers. It indicates the different aspects of the sentence encoded in different hops. An interesting finding is that those large probability points in the lower-level layers are easier to appear when a noun or pronoun is fed into the decoder. For example, when we feed \textit{sri} into the decoder to generate \textit{lankan}, the first layer is obviously more important in hop 2. Similar cases can be observed for words \textit{geneva} and \textit{this} in hop 3, where the word embedding layer has a larger weight. This is reasonable because most of these words have specific meanings and do not need high-level representations for modeling large context in disambiguation.
	
	\section{Related Work}
	Training neural networks with multiple stacked layers is challenging. It has been observed that introducing direct connections between layers can drastically improve the performance of deep neural models. Methods include highway networks \cite{srivastava2015highway}, residual connections \cite{he2016deep}, dense connections \cite{huang2017densely}, and fast-forward connections \cite{zhou2016deep}. E.g., in machine translation, residual networks have been a popular way to address the issue due to its simplicity \cite{wu2016google,gehring2017aconv,gehring2017convs2s,vaswani2017attention}. Another related study is \newcite{wang2017deep}. They introduce linear associative units to reduce the length of gradient propagation in recurrent neural networks (RNNs), and demonstrate promising improvements on their RNN-based NMT systems. But previous studies all focus on using the top-level sentence representation for prediction, and ignore the access to the representations encoded in lower-level layers.
	
	The next obvious step is toward models that make full use of all stacked layers for prediction (call it representation fusion). Some research groups have been aware of this and explored solutions. E.g., \newcite{gehring2017convs2s} find that NMT systems can benefit from shortcut connections from the source word embedding layer to the attention layer.
	Perhaps the most related work is \newcite{xiong2017multi}. They propose a multi-channel encoder (MCE) which uses an external memory module \cite{graves2014neural} to compose word embeddings and hidden states in their RNN-based encoder. But this model is applied to the shallow network on the encoder side. In this work we instead propose a more general method to do representation fusion in either the encoder, the decoder, or both. In addition, we present a 2D representation of sentence which has not been well studied in machine translation.
	
	\section{Conclusion}
	We have proposed a multi-layer representation fusion approach that densely connects all the stacked layers to a fusion layer for encoder or/and decoder in NMT. Also, we have developed three fusion functions to learn a better representation of sentence.
	Experimental results on German-English and Chinese-English translation show that our method outperforms the strong Transformer baseline significantly, and achieves new state-of-the-art on the IWSLT German-English MT task. More interestingly, it is observed that our approach has a regularizing effect and reduces the risk of over-fitting.
	
	\section*{Acknowledgements}
	This work was supported in part by the National Science Foundation of China (No. 61672138, 61432013 and 61572120),  the Fundamental Research Funds for the Central Universities. The authors would like to thank anonymous reviewers and Chunliang Zhang for their comments.
	
	\bibliographystyle{acl}
	\bibliography{coling2018}

\begin{thebibliography}{}

\bibitem[\protect\citename{Ba \bgroup et al.\egroup }2016]{ba16layer}
Lei~Jimmy Ba, Ryan Kiros, and Geoffrey~E. Hinton.
\newblock 2016.
\newblock Layer normalization.
\newblock {\em CoRR}, abs/1607.06450.

\bibitem[\protect\citename{Bahdanau \bgroup et al.\egroup
  }2015]{bahdanau2014neural}
Dzmitry Bahdanau, Kyunghyun Cho, and Yoshua Bengio.
\newblock 2015.
\newblock Neural machine translation by jointly learning to align and
  translate.
\newblock In {\em In Proceedings of the 3rd International Conference on
  Learning Representations}.

\bibitem[\protect\citename{Bahdanau \bgroup et al.\egroup
  }2016]{bahdanau+al-2016-actorcritic}
Dzmitry Bahdanau, Philemon Brakel, Kelvin Xu, Anirudh Goyal, Ryan Lowe, Joelle
  Pineau, Aaron Courville, and Yoshua Bengio.
\newblock 2016.
\newblock An actor-critic algorithm for sequence prediction.
\newblock {\em arXiv e-prints}, abs/1607.07086, July.

\bibitem[\protect\citename{Cettolo \bgroup et al.\egroup
  }2014]{cettolo2014report}
Mauro Cettolo, Jan Niehues, Sebastian St{\"u}ker, Luisa Bentivogli, and
  Marcello Federico.
\newblock 2014.
\newblock Report on the 11th iwslt evaluation campaign, iwslt 2014.
\newblock In {\em Proceedings of the International Workshop on Spoken Language
  Translation, Hanoi, Vietnam}.

\bibitem[\protect\citename{Denkowski and Neubig}2017]{denkowski2017stronger}
Michael Denkowski and Graham Neubig.
\newblock 2017.
\newblock Stronger baselines for trustable results in neural machine
  translation.
\newblock In {\em Proceedings of the First Workshop on Neural Machine
  Translation}, pages 18--27. Association for Computational Linguistics.

\bibitem[\protect\citename{Edunov \bgroup et al.\egroup
  }2017]{edunov2017classical}
Sergey Edunov, Myle Ott, Michael Auli, David Grangier, and Marc'Aurelio
  Ranzato.
\newblock 2017.
\newblock Classical structured prediction losses for sequence to sequence
  learning.
\newblock {\em arXiv preprint arXiv:1711.04956}.

\bibitem[\protect\citename{Gehring \bgroup et al.\egroup
  }2017a]{gehring2017aconv}
Jonas Gehring, Michael Auli, David Grangier, and Yann Dauphin.
\newblock 2017a.
\newblock A convolutional encoder model for neural machine translation.
\newblock In {\em Proceedings of the 55th Annual Meeting of the Association for
  Computational Linguistics (Volume 1: Long Papers)}, pages 123--135.
  Association for Computational Linguistics.

\bibitem[\protect\citename{Gehring \bgroup et al.\egroup
  }2017b]{gehring2017convs2s}
Jonas Gehring, Michael Auli, David Grangier, Denis Yarats, and Yann~N Dauphin.
\newblock 2017b.
\newblock {Convolutional Sequence to Sequence Learning}.
\newblock {\em ArXiv e-prints}, May.

\bibitem[\protect\citename{Graves \bgroup et al.\egroup
  }2014]{graves2014neural}
Alex Graves, Greg Wayne, and Ivo Danihelka.
\newblock 2014.
\newblock Neural turing machines.
\newblock {\em arXiv preprint arXiv:1410.5401}.

\bibitem[\protect\citename{He \bgroup et al.\egroup }2016a]{he2016deep}
Kaiming He, Xiangyu Zhang, Shaoqing Ren, and Jian Sun.
\newblock 2016a.
\newblock Deep residual learning for image recognition.
\newblock In {\em Proceedings of the IEEE conference on computer vision and
  pattern recognition}, pages 770--778.

\bibitem[\protect\citename{He \bgroup et al.\egroup }2016b]{he2016identity}
Kaiming He, Xiangyu Zhang, Shaoqing Ren, and Jian Sun.
\newblock 2016b.
\newblock Identity mappings in deep residual networks.
\newblock In {\em European Conference on Computer Vision}, pages 630--645.
  Springer.

\bibitem[\protect\citename{Huang \bgroup et al.\egroup
  }2017a]{huang2017densely}
Gao Huang, Zhuang Liu, Kilian~Q Weinberger, and Laurens van~der Maaten.
\newblock 2017a.
\newblock Densely connected convolutional networks.
\newblock In {\em Proceedings of the IEEE conference on computer vision and
  pattern recognition}, volume~1, page~3.

\bibitem[\protect\citename{Huang \bgroup et al.\egroup }2017b]{huang2017neural}
Po-Sen Huang, Chong Wang, Dengyong Zhou, and Li~Deng.
\newblock 2017b.
\newblock Neural phrase-based machine translation.
\newblock {\em arXiv preprint arXiv:1706.05565}.

\bibitem[\protect\citename{Kingma and Ba}2014]{kingma2014adam}
Diederik~P Kingma and Jimmy Ba.
\newblock 2014.
\newblock Adam: A method for stochastic optimization.
\newblock {\em arXiv preprint arXiv:1412.6980}.

\bibitem[\protect\citename{Lin \bgroup et al.\egroup }2017]{lin2017structured}
Zhouhan Lin, Minwei Feng, Cicero~Nogueira dos Santos, Mo~Yu, Bing Xiang, Bowen
  Zhou, and Yoshua Bengio.
\newblock 2017.
\newblock A structured self-attentive sentence embedding.

\bibitem[\protect\citename{Liu \bgroup et al.\egroup }2018]{liu2018path}
Shu Liu, Lu~Qi, Haifang Qin, Jianping Shi, and Jiaya Jia.
\newblock 2018.
\newblock Path aggregation network for instance segmentation.
\newblock {\em arXiv preprint arXiv:1803.01534}.

\bibitem[\protect\citename{Luong \bgroup et al.\egroup }2015]{luong2015address}
Thang Luong, Ilya Sutskever, Quoc Le, Oriol Vinyals, and Wojciech Zaremba.
\newblock 2015.
\newblock Addressing the rare word problem in neural machine translation.
\newblock In {\em Proceedings of the 53rd Annual Meeting of the Association for
  Computational Linguistics and the 7th International Joint Conference on
  Natural Language Processing (Volume 1: Long Papers)}, pages 11--19.
  Association for Computational Linguistics.

\bibitem[\protect\citename{Luong \bgroup et al.\egroup }2017]{luong17}
Minh{-}Thang Luong, Eugene Brevdo, and Rui Zhao.
\newblock 2017.
\newblock Neural machine translation (seq2seq) tutorial.
\newblock {\em https://github.com/tensorflow/nmt}.

\bibitem[\protect\citename{Ranzato \bgroup et al.\egroup
  }2015]{ranzato2015sequence}
Marc'Aurelio Ranzato, Sumit Chopra, Michael Auli, and Wojciech Zaremba.
\newblock 2015.
\newblock Sequence level training with recurrent neural networks.
\newblock {\em arXiv preprint arXiv:1511.06732}.

\bibitem[\protect\citename{Sennrich \bgroup et al.\egroup
  }2016]{sennrich2015neural}
Rico Sennrich, Barry Haddow, and Alexandra Birch.
\newblock 2016.
\newblock Neural machine translation of rare words with subword units.
\newblock In {\em Proceedings of the 54th Annual Meeting of the Association for
  Computational Linguistics, {ACL} 2016, August 7-12, 2016, Berlin, Germany,
  Volume 1: Long Papers}.

\bibitem[\protect\citename{Sennrich \bgroup et al.\egroup
  }2017]{sennrich-EtAl:2017:EACLDemo}
Rico Sennrich, Orhan Firat, Kyunghyun Cho, Alexandra Birch, Barry Haddow,
  Julian Hitschler, Marcin Junczys-Dowmunt, Samuel L\"{a}ubli, Antonio~Valerio
  Miceli~Barone, Jozef Mokry, and Maria Nadejde.
\newblock 2017.
\newblock Nematus: a toolkit for neural machine translation.
\newblock In {\em Proceedings of the Software Demonstrations of the 15th
  Conference of the European Chapter of the Association for Computational
  Linguistics}, pages 65--68, Valencia, Spain, April. Association for
  Computational Linguistics.

\bibitem[\protect\citename{Shen \bgroup et al.\egroup }2018]{shen2017disan}
Tao Shen, Tianyi Zhou, Guodong Long, Jing Jiang, Shirui Pan, and Chengqi Zhang.
\newblock 2018.
\newblock Disan: Directional self-attention network for rnn/cnn-free language
  understanding.
\newblock In {\em AAAI Conference on Artificial Intelligence}.

\bibitem[\protect\citename{Srivastava \bgroup et al.\egroup
  }2015]{srivastava2015highway}
Rupesh~K Srivastava, Klaus Greff, and J\"{u}rgen Schmidhuber.
\newblock 2015.
\newblock Training very deep networks.
\newblock In C.~Cortes, N.~D. Lawrence, D.~D. Lee, M.~Sugiyama, and R.~Garnett,
  editors, {\em Advances in Neural Information Processing Systems 28}, pages
  2377--2385. Curran Associates, Inc.

\bibitem[\protect\citename{Vaswani \bgroup et al.\egroup
  }2017]{vaswani2017attention}
Ashish Vaswani, Noam Shazeer, Niki Parmar, Jakob Uszkoreit, Llion Jones,
  Aidan~N Gomez, {\L}ukasz Kaiser, and Illia Polosukhin.
\newblock 2017.
\newblock Attention is all you need.
\newblock In {\em Advances in Neural Information Processing Systems}, pages
  6000--6010.

\bibitem[\protect\citename{Wang \bgroup et al.\egroup }2017]{wang2017deep}
Mingxuan Wang, Zhengdong Lu, Jie Zhou, and Qun Liu.
\newblock 2017.
\newblock Deep neural machine translation with linear associative unit.
\newblock In {\em Proceedings of the 55th Annual Meeting of the Association for
  Computational Linguistics, {ACL} 2017, Vancouver, Canada, July 30 - August 4,
  Volume 1: Long Papers}, pages 136--145.

\bibitem[\protect\citename{Wiseman and Rush}2016]{wiseman2016sequence}
Sam Wiseman and Alexander~M. Rush.
\newblock 2016.
\newblock Sequence-to-sequence learning as beam-search optimization.
\newblock In {\em Proceedings of the 2016 Conference on Empirical Methods in
  Natural Language Processing, {EMNLP} 2016, Austin, Texas, USA, November 1-4,
  2016}, pages 1296--1306.

\bibitem[\protect\citename{Wu \bgroup et al.\egroup }2016]{wu2016google}
Yonghui Wu, Mike Schuster, Zhifeng Chen, Quoc~V Le, Mohammad Norouzi, Wolfgang
  Macherey, Maxim Krikun, Yuan Cao, Qin Gao, Klaus Macherey, et~al.
\newblock 2016.
\newblock Google's neural machine translation system: Bridging the gap between
  human and machine translation.
\newblock {\em arXiv preprint arXiv:1609.08144}.

\bibitem[\protect\citename{Xiao \bgroup et al.\egroup }2012]{xiao2012niutrans}
Tong Xiao, Jingbo Zhu, Hao Zhang, and Qiang Li.
\newblock 2012.
\newblock Niutrans: an open source toolkit for phrase-based and syntax-based
  machine translation.
\newblock In {\em Proceedings of the ACL 2012 System Demonstrations}, pages
  19--24. Association for Computational Linguistics.

\bibitem[\protect\citename{Xiong \bgroup et al.\egroup }2017]{xiong2017multi}
Hao Xiong, Zhongjun He, Xiaoguang Hu, and Hua Wu.
\newblock 2017.
\newblock Multi-channel encoder for neural machine translation.
\newblock {\em arXiv preprint arXiv:1712.02109}.

\bibitem[\protect\citename{Zhou \bgroup et al.\egroup }2016]{zhou2016deep}
Jie Zhou, Ying Cao, Xuguang Wang, Peng Li, and Wei Xu.
\newblock 2016.
\newblock Deep recurrent models with fast-forward connections for neural
  machine translation.
\newblock {\em Transactions of the Association of Computational Linguistics},
  4:371--383.

\end{thebibliography}

\end{document}